\documentclass[letterpaper]{article} 
\usepackage{aaai25}  
\nocopyright

\newif\ifCheckUnusedRefs

\usepackage{times}  
\usepackage{helvet}  
\usepackage{courier}  
\usepackage[hyphens]{url}  
\usepackage{graphicx} 
\usepackage{natbib}  
\usepackage{caption} 
\usepackage[hidelinks]{hyperref}

\usepackage{adjustbox}
\usepackage{placeins}

\usepackage[utf8]{inputenc}
\usepackage{bm}         
\usepackage{amssymb}
\usepackage{amsthm}
\usepackage{amsmath}
\usepackage{enumitem}
\usepackage{caption}
\usepackage{subcaption}
\usepackage{centernot}
\usepackage{complexity}
\usepackage{mathrsfs}   
\usepackage{tabularx}   
\usepackage{multirow}   
\usepackage{array}      
\usepackage{booktabs}   
\usepackage{colortbl}   
\usepackage{xargs}      
\usepackage{tikz}
\usepackage{tikz-cd}
\usepackage{url}
\usepackage{makecell}   
\usepackage[ruled,noline,linesnumbered,noend]{algorithm2e} 
\usepackage{float}
\usepackage{etoc}
\usepackage{todonotes}
\ifCheckUnusedRefs
    \usepackage{refcheck}   
\fi

\usetikzlibrary{positioning, calc}

\setlist[enumerate]{label=(\arabic*)}

\newcommand{\domain}{\mathcal{D}}

\newcommand{\predicates}{\mathcal{P}}

\newcommand{\objects}{\mathcal{O}}
\newcommand{\schemata}{\mathcal{A}}
\newcommand{\goal}{\mathcal{G}}

\newcommand{\training}{\mathcal{T}_\text{train}}
\newcommand{\testing}{\mathcal{T}_\text{test}}

\newcommand{\graph}{\mathbf{G}}
\newcommand{\nodes}{\mathbf{V}}
\newcommand{\edges}{\mathbf{E}}

\newcommand{\neighbour}{\mathcal{N}}

\newcommand{\gnn}{\mathrm{GNN}}

\newcommand{\features}{\mathcal{F}}

\newcommand{\ob}{\texttt{ob}}
\newcommand{\ap}{\texttt{ap}}
\newcommand{\ag}{\texttt{ag}}
\newcommand{\ug}{\texttt{ug}}

\newclass{\N}{N}
\newclass{\CountingLogic}{C}
\newclass{\coNTIME}{coNTIME}
\newclass{\coNSPACE}{coNSPACE}
\newclass{\coNPSPACE}{coNPSPACE}
\newclass{\EXPTIME}{EXPTIME}
\newclass{\NEXPTIME}{NEXPTIME}
\newclass{\coNEXPTIME}{coNEXPTIME}
\newclass{\NEXPSPACE}{NEXPSPACE}
\newclass{\coNEXPSPACE}{coNEXPSPACE}
\newclass{\ASPACE}{ASPACE}
\newclass{\ATIME}{ATIME}
\newclass{\APSPACE}{APSPACE}
\newclass{\AEXPTIME}{AEXPTIME}
\newclass{\AEXPSPACE}{AEXPSPACE}

\theoremstyle{definition}

\def\N{\mathbb{N}}
\def\R{\mathbb{R}}

\def\g{\gamma}

\renewcommand{\phi}{\varphi}

\newcommand{\gen}[1]{\left< #1 \right>}

\newcommand{\set}[1]{\left\{ #1 \right\}}

\newcommand{\brk}[1]{\left[ #1 \right]}

\newcommand{\zerocell}[1]{-}

\newcolumntype{Y}{>{\raggedleft\arraybackslash}X}

\definecolor{caribbeangreen}{rgb}{0.0, 0.8, 0.6}
\definecolor{brilliantlavender}{rgb}{0.96, 0.73, 1.0}
\definecolor{amethyst}{rgb}{0.6, 0.4, 0.8}
\definecolor{ao(english)}{rgb}{0.0, 0.5, 0.0}
\definecolor{arylideyellow}{rgb}{0.91, 0.84, 0.42}
\definecolor{asparagus}{rgb}{0.53, 0.66, 0.42}
\definecolor{aquamarine}{rgb}{0.5, 1.0, 0.83}
\definecolor{babyblue}{rgb}{0.54, 0.81, 0.94}
\definecolor{fwtchanged}{rgb}{0.3, 0.3, 0.7}
\definecolor{rosewood}{rgb}{0.4, 0.0, 0.04}
\definecolor{oldmauve}{rgb}{0.4, 0.19, 0.28}
\definecolor{myrtle}{rgb}{0.13, 0.26, 0.12}
\definecolor{magenta(dye)}{rgb}{0.79, 0.08, 0.48}

\definecolor{plta}{rgb}{0.12, 0.47, 0.71}
\definecolor{pltb}{rgb}{   1, 0.5, 0.05}
\definecolor{pltc}{rgb}{0.17, 0.63, 0.17}
\definecolor{pltd}{rgb}{0.84, 0.15, 0.16}

\newcommand{\svr}{\mathrm{SVR}}
\newcommand{\gpr}{\mathrm{GPR}}
\newcommand{\svm}{\mathrm{SVM}}
\newcommand{\gpc}{\mathrm{GPC}}
\newcommand{\lp}{\mathrm{LP}}
\newcommand{\mip}{\mathrm{MIP}}

\definecolor{caribbeangreen}{rgb}{0.0, 0.8, 0.6}
\definecolor{brilliantlavender}{rgb}{0.96, 0.73, 1.0}
\definecolor{amethyst}{rgb}{0.6, 0.4, 0.8}
\definecolor{ao(english)}{rgb}{0.0, 0.5, 0.0}
\definecolor{arylideyellow}{rgb}{0.91, 0.84, 0.42}
\definecolor{asparagus}{rgb}{0.53, 0.66, 0.42}
\definecolor{aquamarine}{rgb}{0.5, 1.0, 0.83}
\definecolor{babyblue}{rgb}{0.54, 0.81, 0.94}
\definecolor{fwtchanged}{rgb}{0.3, 0.3, 0.7}
\definecolor{rosewood}{rgb}{0.4, 0.0, 0.04}
\definecolor{oldmauve}{rgb}{0.4, 0.19, 0.28}
\definecolor{myrtle}{rgb}{0.13, 0.26, 0.12}
\definecolor{magenta(dye)}{rgb}{0.79, 0.08, 0.48}

\definecolor{plta}{rgb}{0.12, 0.47, 0.71}
\definecolor{pltb}{rgb}{   1, 0.5, 0.05}
\definecolor{pltc}{rgb}{0.17, 0.63, 0.17}
\definecolor{pltd}{rgb}{0.84, 0.15, 0.16}

\newcommand{\sssection}[1]{\subsubsection{#1}}

\newcommand{\colourdc}{blue}
  
\newcommand{\todocustom}[3]{\todo[linecolor=#2,backgroundcolor=#2!25,bordercolor=#2,size=\tiny,#3]{#1}}  
\newcommandx{\nbdc}[2][1=]{\todocustom{#2}{\colourdc}{#1}}

\def\N{\mathbb{N}}
\def\R{\mathbb{R}}

\def\g{\gamma}

\renewcommand{\phi}{\varphi}

\newcommand{\range}[1]{\left[\!\left[ #1 \right]\!\right]}

\renewcommand{\objects}{\mathcal{O}}
\renewcommand{\predicates}{\Sigma_{p}}
\newcommand{\functions}{\Sigma_{f}}
\renewcommand{\schemata}{\mathcal{A}}
\renewcommand{\goal}{\mathcal{G}}

\newcommand{\arity}{\mathrm{arity}}

\newcommand{\graphFont}[1]{\mathbf{#1}}
\renewcommand{\neighbour}{\graphFont{N}}
\newcommand{\featCat}{\graphFont{F}_{\text{cat}}}
\newcommand{\featCon}{\graphFont{F}_{\text{con}}}
\newcommand{\featEdge}{\graphFont{L}}
\renewcommand{\graph}{\graphFont{G}}
\renewcommand{\nodes}{\graphFont{V}}
\renewcommand{\edges}{\graphFont{E}}

\newcommand{\nodeCat}{\Sigma_{\text{V}}}
\newcommand{\edgeCat}{\Sigma_{\text{E}}}

\title{Graph Learning for Planning:\\The Story Thus Far and Open Challenges}
\author{
    $^{1,2}$Dillon Z. Chen, $^2$Mingyu Hao, $^{1,2}$Sylvie Thi\'ebaux, $^2$Felipe Trevizan\\
}
\affiliations{
    $^1$LAAS-CNRS, University of Toulouse \\
    $^2$The Australian National University \\
    firstName.lastName@anu.edu.au
}

\begin{document}

\maketitle


\begin{abstract}
Graph learning is naturally well suited for use in planning due to its ability to exploit relational structures exhibited in planning domains and to take as input planning instances with arbitrary number of objects. In this paper, we study the usage of graph learning for planning thus far by studying the theoretical and empirical effects on learning and planning performance of (1) graph representations of planning tasks, (2) graph learning architectures, and (3) optimisation formulations for learning. Our studies accumulate in the GOOSE framework which learns domain knowledge from small planning tasks in order to scale up to much larger planning tasks. In this paper, we also highlight and propose the 5 open challenges in the general Learning for Planning field that we believe need to be addressed for advancing the state-of-the-art.
\end{abstract}

\section{Introduction}
Learning for Planning (L4P) has gained significant interest in recent years due to advancements of machine learning (ML) approaches across various fields, and also because planning is one of the few problems in AI that has been unsolved by deep learning and large models~\cite{valmeekam.etal.2023,valmeekam.etal.2023a,valmeekam.etal.2024}.
An aim of L4P involves designing automated, domain-independent algorithms for learning domain knowledge from small training problems for scaling up planning to problems of arbitrary size~\cite{jimenez.etal.2012,toyer.etal.2018,toyer.etal.2020,dong.etal.2019,shen.etal.2020,karia.srivastava.2021,staahlberg.etal.2022,staahlberg.etal.2023,mao.etal.2023,chen.etal.2024,chen.etal.2024a,wang.thiebaux.2024,hao.etal.2024,chen.thiebaux.2024}.
Planning tasks can exhibit arbitrary numbers of objects and are represented with relational languages which begs for the use of learning approaches that operate on relational structures such as graphs.

\begin{figure}[t]
    \centering
    \newcommand{\yShift}{3.199cm}
    \scalebox{0.98}{
            \trimbox{0.3cm 0.16cm 0.36cm 0.31cm}{
                \input{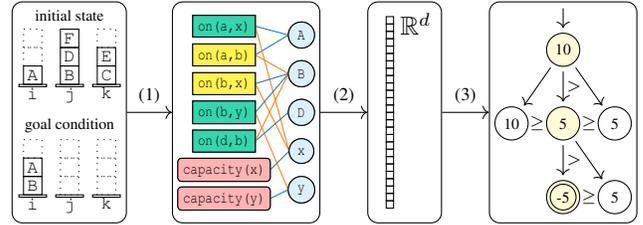}
            }
    }
    \caption{
        Typical graph learning for planning pipeline.
        (1) Planning tasks are represented as graphs.
        (2) Graph learning architectures are used to embed graphs.
        (3) Learning is performed via some optimisation criteria suited for making predictions in planning.
    }
    \label{fig:graph-learning}
\end{figure}

In this paper, we recount the story so far for graph learning for planning.
This will be done by studying the contributions in the literature and of the authors thus far regarding the three main components of graph learning for planning as summarised in Fig.~\ref{fig:graph-learning}: (1) graph representations of planning tasks, (2) graph learning architectures, and (3) optimisation formulations for learning.
On the experimental side, we present the GOOSE\footnote{\textbf{G}raphs \textbf{O}ptimised f\textbf{O}r \textbf{S}earch \textbf{E}valuation; code available at \url{https://github.com/DillonZChen/goose}} learning framework for leveraging graph learning for planning.

This paper summarises previous~\cite{chen.etal.2024,chen.etal.2024a,hao.etal.2024,chen.thiebaux.2024} and current theoretical and experimental work by the authors.
The paper is structured by introducing the general L4P problem setup in Section~\ref{sec:l4p} and the formal background and definitions in Section~\ref{sec:background}.
Next, we highlight contributions of the authors in graph learning for planning, summarised as follows:
\begin{enumerate}[label=(\arabic*)]
    \item We provide a taxonomy and theoretical comparison of graph representations of planning tasks and their expressive power with graph learning approaches. (Section~\ref{sec:graph})
    \item We introduce a classical ML approach which outperforms deep learning methods by several orders of magnitude for planning over various metrics. (Section~\ref{sec:models})
    \item We present optimisers better suited for learning for planning by representing heuristic functions as rankings of states rather than cost-to-go estimates. (Section~\ref{sec:optimisation})
\end{enumerate}

Section~\ref{sec:experiments} quantifies the contributions of our work by showcasing experimental results.
We conclude by highlighting open problems and challenges for the general L4P field in Section~\ref{sec:challenges}, and avenues for future work in the field.

\section{Learning for Planning}\label{sec:l4p}
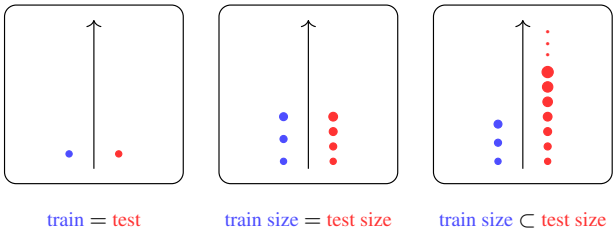
\begin{figure}
    \centering
    \scalebox{1.1}{\begin{tikzpicture}
    \newcommand{\trainColour}{blue!70}
    \newcommand{\testColour}{red!80}
    \newcommand{\axisHeight}{1.8}
    \newcommand{\nodeSize}{0.075cm}
    \newcommand{\nodeInc}{0.01cm}
    \newcommand{\gap}{0.3}
    \newcommand{\borderSize}{\axisHeight*1.2cm}
    \newcommand{\segmentC}{1.2}
    \newcommand{\textLoc}{-0.7*\borderSize}
    \tikzset{
        trainNode/.style={
                inner sep=0,
                circle,
                draw=\trainColour,
                fill=\trainColour,
                minimum width=\nodeSize,
            },
        testNode/.style={
                inner sep=0,
                circle,
                draw=\testColour,
                fill=\testColour,
                minimum width=\nodeSize,
            },
        border/.style={
                draw,
                rectangle,
                rounded corners,
                minimum width=\borderSize,
                minimum height=\borderSize,
            },
    }
    \newcommand{\border}[1]{
        \node[border] at (0,0) {};
        \node[font=\scriptsize] at (0,\textLoc) {#1}
    }
    \newcommand{\axis}{
        \draw[->] (0,-\axisHeight/2) -- (0,\axisHeight/2) 
    }
    \begin{scope}[xshift=-\segmentC*\borderSize]
        \border{{\color{\trainColour}train} $=$ {\color{\testColour}test}};
        \axis;
        \node[trainNode] at (-\gap,-.4*\axisHeight) {};
        \node[testNode]  at ( \gap,-.4*\axisHeight) {};
    \end{scope}
    \begin{scope}
        \border{{\color{\trainColour}train size} $=$ {\color{\testColour}test size}};
        \axis;
        \node[trainNode] at (-\gap,-.45*\axisHeight) {};
        \node[trainNode,minimum size=\nodeSize+1*\nodeInc] at (-\gap,-.3*\axisHeight) {};
        \node[trainNode,minimum size=\nodeSize+2*\nodeInc] at (-\gap,-.15*\axisHeight) {};
        \node[testNode]  at ( \gap,-.45*\axisHeight) {};
        \node[testNode,minimum size=\nodeSize+1*\nodeInc]  at ( \gap,-.35*\axisHeight) {};
        \node[testNode,minimum size=\nodeSize+2*\nodeInc]  at ( \gap,-.25*\axisHeight) {};
        \node[testNode,minimum size=\nodeSize+3*\nodeInc]  at ( \gap,-.15*\axisHeight) {};
    \end{scope}
    \begin{scope}[xshift=\segmentC*\borderSize]
        \border{{\color{\trainColour}train size} $\subset$ {\color{\testColour}test size}};
        \axis;
        \node[trainNode] at (-\gap,-.45*\axisHeight) {};
        \node[trainNode,minimum size=\nodeSize+1*\nodeInc] at (-\gap,-.325*\axisHeight) {};
        \node[trainNode,minimum size=\nodeSize+2*\nodeInc] at (-\gap,-.2*\axisHeight) {};
        \node[testNode]  at ( \gap,-.45*\axisHeight) {};
        \node[testNode,minimum size=\nodeSize+1*\nodeInc]  at ( \gap,-.35*\axisHeight) {};
        \node[testNode,minimum size=\nodeSize+2*\nodeInc]  at ( \gap,-.25*\axisHeight) {};
        \node[testNode,minimum size=\nodeSize+3*\nodeInc]  at ( \gap,-.15*\axisHeight) {};
        \node[testNode,minimum size=\nodeSize+4*\nodeInc]  at ( \gap,-.05*\axisHeight) {};
        \node[testNode,minimum size=\nodeSize+5*\nodeInc]  at ( \gap, .05*\axisHeight) {};
        \node[testNode,minimum size=\nodeSize+6*\nodeInc]  at ( \gap, .15*\axisHeight) {};
        \node at ( \gap, .4*\axisHeight) {\color{\testColour}$\vdots$};
    \end{scope}
\end{tikzpicture}}
    \caption{
        Visualisations of generalisation setups for planning and RL.
        The axis represents the number of objects in a planning task.
        Standalone planning and RL commonly follow the task-specific setup (left), with some RL approaches generalising across similar-sized tasks (middle).
        L4P approaches primarily focuses on generalisation across arbitrary numbers of objects (right).
    }
    \label{fig:generalisation}
\end{figure}

Here we briefly introduce and describe the general Learning for Planning (L4P) problem statement.
The problem setup of L4P involves learning knowledge from a set of training problems in either a supervised, unsupervised, or reinforcement learning fashion that may be helpful for planning.
This is in contrast to planning for each individual problem from scratch in typical planning setups~\cite{geffner.bonet.2013}, and learning to solve a specific problem instance from rewards typical in task-specific reinforcement learning~\cite{sutton.barto.1998} or across different initial states~\cite{ferber.etal.2022}.
Furthermore, as illustrated in Fig.~\ref{fig:generalisation}, L4P aims to generalise across an arbitrary number of objects in the planning domain, as opposed to transfer learning across similar-sized tasks or task-specific reinforcement learning.
The primary objective of L4P is that by performing either few-shot training or bootstrapping, we can scale up to much larger planning tasks which domain-independent planners cannot handle due to no free lunch.

We define an L4P problem as any tuple $\gen{\domain, \training, \testing}$ where $\domain$ is a domain characterising a class of planning tasks, $\training$ is a finite set of training tasks from $\domain$, and $\testing$ is a (possibly infinite) set of testing tasks from $\domain$.
An L4P solution is defined as a set of plans corresponding to problems in $\testing$.

L4P opens a rich array of possibilities for how one may come up with algorithms for solving the problem, and methods for evaluating such algorithms.
We can characterise an L4P algorithm as an algorithm with two components:
\begin{enumerate}[label=(\arabic*)]
    \item a \textbf{learner module}, whose input is $\domain$ and $\training$ and output is a \emph{knowledge} artifact, and
    \item a \textbf{planner module}, whose input is a task from $\testing$ and the \emph{knowledge} artifact, and output is a plan.
\end{enumerate}

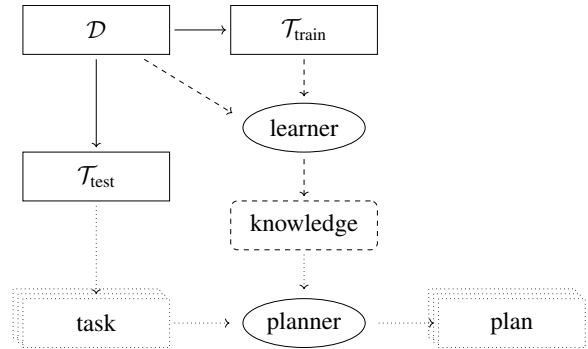
\begin{figure}
    \centering
    \LARGE
    \scalebox{0.65}{\begin{tikzpicture}
    \newcommand{\xshift}{4.25cm}
    \newcommand{\yshift}{-2cm}
    \newcommand{\trainY}{2.5cm}

    \newcommand{\ovalX}{1.25cm}
    \newcommand{\ovalY}{0.5cm}
    \newcommand{\ovalDim}{(1.25cm and 0.5cm)}

    \tikzset{
        align=center,
        outer sep=1mm,  
        rect/.style={
                rectangle,
                draw,
                minimum width=3cm,
                minimum height=1cm,
            },
        model/.style={
                minimum width=10cm,
                minimum height=2cm,
            },
        oval/.style={
                draw,
            },
        io/.style={
                dotted,
            },
        knowledge/.style={
                rounded corners,
                dashed,
            },
    }

    \node[rect] (d) at (0, 0) {$\domain$};

    \node[rect] (train) at (\xshift, 0) {$\training$};

    \node[rect] (test) at (0, 1.5*\yshift) {$\testing$};

    \node (learner) at (\xshift, \yshift) {learner};
    \draw[oval] (learner) ellipse \ovalDim;

    \node[rect,knowledge] (knowledge) at (\xshift, 2*\yshift) {knowledge};

    \node (planner) at (\xshift, 3*\yshift) {planner};
    \draw[oval] (planner) ellipse \ovalDim;

    \newcommand{\nudge}{1mm}
    \node[rect,io] at (0-2*\nudge, 3*\yshift+2*\nudge) {};
    \node[rect,io,fill=white] at (0-\nudge, 3*\yshift+\nudge) {};
    \node[rect,io,fill=white] (task) at (0, 3*\yshift) {task};

    \node[rect,io] at (2*\xshift-2*\nudge, 3*\yshift+2*\nudge) {};
    \node[rect,io,fill=white] at (2*\xshift-\nudge, 3*\yshift+\nudge) {};
    \node[rect,io,fill=white] (plan) at (2*\xshift, 3*\yshift) {plan};

    \draw[->] (d) -- (train);
    \draw[->] (d) -- (test);

    \draw[->,dashed] (d)     -- ($(learner.west) + (-5mm, 3mm)$);
    \draw[->,dashed] (train) -- ($(learner.north) + (0, 2mm)$);
    \draw[->,dashed] ($(learner.south) + (0, -2mm)$) -- (knowledge);

    \draw[->,dotted] (test) -- (task);
    \draw[->,dotted] (knowledge) -- ($(planner.north) + (0, 2mm)$);
    \draw[->,dotted] (task) -- ($(planner.west) + (-5mm, 0)$);
    \draw[->,dotted] ($(planner.east) + (5mm, 0)$) -- (plan);
\end{tikzpicture}}
    \caption{
        The general Learning for Planning (L4P) setup.
    }
    \label{fig:l4p}
\end{figure}

Fig.~\ref{fig:l4p} illustrates the general L4P setup and pipeline.
The learner module learns from the training tasks and uses the domain as prior knowledge to produce a knowledge artifact, which can then be used by a planner module to plan for arbitrary tasks from $\testing$.
Common knowledge artifacts include heuristic or value functions, and global policies.

\section{Background}\label{sec:background}
In this section, we provide the technical background and formalisms of planning used in works described in the paper.
However, they may be skipped upon first read as they are not entirely necessary for a majority of the paper.

\subsection{Planning Task}
Let $\range{n}$ denote the set of integers $\set{1,\ldots,n}$.
A planning task can be understood as a state transition model~\cite{geffner.bonet.2013} given by a tuple $\Pi = \gen{S, A, s_0, G}$ where $S$ is a set of states, $A$ a set of actions, $s_0 \in S$ an initial state, and $G \subseteq S$ a set of goal states.
Each action $a \in A$ is a function $a: S \rightarrow S \cup \set{\bot}$ where $a(s) = \bot$ if $a$ is not applicable in $s$, and $a(s) \in S$ is the successor state when $a$ is applied to $s$.
A solution for a planning task is a plan: a sequence of actions $\pi = a_1, \ldots, a_n$ where $s_i = a_i(s_{i-1}) \not= \bot$ for $i \in \range{n}$ and $s_n \in G$.
A state $s$ in a planning task $\Pi$ induces a new planning task $\Pi' = \gen{S, A, s, G}$.
A planning task is solvable if there exists at least one plan.

\subsection{Planning Representation}
We now describe the numeric planning formalism from PDDL2.1~\cite{fox.long.2003}.
Numeric planning encapsulates classical planning which is a compact, lifted representation of a planning task with the use of predicate logic and relational numeric variables.
More specifically, a numeric planning task is a tuple $\Pi = \gen{\objects, \predicates, \functions, \schemata, s_0, \goal}$, where $\objects$ denotes a set of objects, $\predicates$/$\functions$ a set of predicate/function symbols, $\schemata$ a set of action schemata, $s_0$ the initial state, and $\goal$ now the goal condition.
Details of the representation of actions in a planning task induced by grounding action schemata from $\schemata$ are not required for understanding the paper.
We instead focus on the representation of states and the goal condition.

\sssection{States}
Each symbol $\sigma \in \predicates \cup \functions$ is associated with an arity $\arity(\sigma)\in\N \cup \set{0}$.
Predicates and functions take the form $p(x_1,\ldots,x_{n_p})$ and $f(x_1,\ldots,x_{n_f})$ respectively, where the $x_i$ denotes the $i$th argument.
Propositional and numeric variables are defined by substituting objects into predicate and function variables.
%
%
%
A state $s$ is an assignment of values in $\set{\top, \bot}$ (resp. $\R$) to all possible propositional (resp. numeric) variables in a state.
Following the closed world assumption, we can equivalently represent a state as a set of true propositions and numeric assignments.
%

\sssection{Goal Condition}
A propositional condition is a positive (resp. negative) literal $x=\top$ (resp. $x=\bot$) where $x$ is a propositional variable.
A numeric condition has the form $\xi \unrhd 0$ where $\xi$ is an arithmetic expression over numeric variables and $\unrhd \in \set{\geq, >, =}$.
%
%
The goal condition $\goal$ is a set of propositional and numeric conditions which we denote $\goal_p$ and $\goal_n$, respectively.
A state $s$ satisfies the goal condition $\goal$ if $s$ satisfies all its conditions.

\sssection{Domain}
A planning domain is a set of planning tasks which share the same set of predicates, functions and action schemata.
Constant objects are objects that occur in all planning tasks in a domain.

\subsection{Graphs}\label{ssec:graph}
We denote a graph with categorical and continuous node features and edge labels by a tuple $\graph = \gen{\nodes, \edges, \featCat, \featCon, \featEdge}$.
We have that $\nodes$ is a set of nodes, $\edges$ a set of edges, $\featCat:\nodes \to \nodeCat$ the categorical node features, $\featCon:\nodes \to \R$ are the continuous node features, and $\featEdge:\edges \to \edgeCat$ the edge labels.
The neighbourhood of a node $u \in \nodes$ in a graph is defined by $\neighbour_{\iota}(u) = \set{v \in \nodes \mid \gen{u,v} \in \edges}$.
The neighbourhood of a node $u \in \nodes$ in a graph with respect to an edge label $\iota$ is defined by $\neighbour_{\iota}(u) = \set{v \in \nodes \mid e=\gen{u,v} \in \edges \land \featEdge(e) = \iota}$.

\section{Graph Representations}\label{sec:graph}
Representations of planning tasks into ML models have a great impact on both learning and planning performance.
Graph representations are particularly well-suited due to their ability to model relational information of planning tasks, as well as their ability to model arbitrarily large planning tasks.
In this section, we unify the graph representations of planning tasks in the learning for planning literature by taxonomising graph definitions.
Next, we summarise theoretical expressivity results concerning such graphs through the lens of their ability to distinguish planning tasks in conjunction with message passing graph neural networks (MPNNs)~\cite{gilmer.etal.2017}.

\subsection{Graph Taxonomy}
The use of predicate logic in planning representations makes it natural for several relational or graph representations to arise from planning tasks.
Indeed, early graph representations for planning tasks were geared towards constructing algorithms or heuristic functions for planners, starting with the planning graph which was a vital component of various early planners~\cite{blum.furst.1997,hoffmann.nebel.2001,gerevini.serina.2002}, to the usage of various different graphs and graph algorithms for computing transformations and heuristics of planning tasks~\cite{helmert.2004,helmert.domshlak.2009}, and also to the use of detecting planning task symmetries~\cite{pochter.etal.2011,shleyfman.etal.2015,sievers.etal.2019}.

On the learning side, graph representations were heavily relied on for representing planning tasks given the unbounded nature of planning task sizes.
ASNets~\cite{toyer.etal.2018,toyer.etal.2020} was the seminal work in this field, which made use of MPNNs for learning policies for probabilistic planning tasks, and was recently extended for numeric planning~\cite{wang.thiebaux.2024}.
Later MPNN approaches for planning focused on learning heuristic or value functions~\cite{shen.etal.2020,staahlberg.etal.2022,staahlberg.etal.2023,chen.etal.2024,chen.etal.2024a,chen.thiebaux.2024}, with exceptions being learning policy rules~\cite{dong.etal.2019}, portfolios~\cite{ma.etal.2020} and detecting object importance~\cite{silver.etal.2021}.

An underlying component of all such learning works is that planning tasks are represented as some form of graph.
However, there is no such work which compares all such definitions and provide a high-level view on similar or different characteristics of such representations.
In our ongoing work, we have identified following classification of many existing graph representations of planning tasks:
\begin{enumerate}
    \item \underline{Grounded Graphs} -- nodes represent all possible ground propositions and actions of planning tasks, and edges are defined from precondition and effect relations of actions. Example graphs include ASNets~\cite{toyer.etal.2018,toyer.etal.2020}, STRIPS-HGN~\cite{shen.etal.2020} and the SLG in GOOSE~\cite{chen.etal.2024}.
    \item \underline{Lifted Graphs with Instantiation Relation (IR)} -- nodes represent task objects and only propositions that are true in the state or the goal condition, and edges are defined by the relations between object instantiations in ground propositions. Example graphs include Muninn~\cite{staahlberg.etal.2022} and the ILG in GOOSE~\cite{chen.etal.2024}.
    \item \underline{Lifted Graphs with Predicate Relation (PR)} -- nodes represent only task objects, and edges are defined by the location of pairs of objects in $n$-ary predicates for $n \geq 2$. This can also be viewed as a line graph of lifted graphs with Object Relation (OR), where nodes represent only task atoms, and edges between atoms sharing an object. Example graphs include the PLOI graph~\cite{silver.etal.2021} and the Object Binary Structure~\cite{horcik.sir.2024}.
\end{enumerate}

\subsection{Graph Hierarchy}\label{ssec:hierarchy}
On top of achieving a taxonomy, we would like to understand whether there are any theoretical or practical relationships between such graphs.
Existing~\cite{chen.etal.2024,chen.etal.2024a} and ongoing work of authors have attempted to understand such relationships through the lens of expressive power.

More specifically, we compare the expressive power of graph representations based on planning tasks they can distinguish when used with MPNNs.
Results primarily bootstrap from existing works measuring the distinguishability of existing general graph learning models by making use of the well-known result that the colour refinement or Weisfeiler-Leman (WL) algorithm subsumes MPNN expressivity~\cite{morris.etal.2019,xu.etal.2019}, with novelty lying in identifying the effect of graph representations of planning tasks.
However, we note that our results differ from graph learning literature results which focus on expressiveness~\cite{morris.etal.2020,morris.etal.2022,zhao.etal.2022,wang.etal.2023,alvarezgonzalez.etal.2024} of \emph{architectures} rather than \emph{representations} as the graph representations are assumed to be fixed in graph learning datasets.
Results are summarised in Fig.~\ref{fig:expressivity} and key takeaways are:
\begin{itemize}
    \item grounded representations implicitly encode instantiation relations and thus are more expressive than lifted representations with IR ({\color{green}{green}} edges)
    \item not encoding planning domain information (predicates and schemata) results in lower expressivity across different graph classes ({\color{red}{red}} edges)
    \item lifted graphs with PR are incomparable to both lifted graphs with IR, and grounded graphs under a weaker notion of expressivity ({\color{blue}{blue}} edges)
\end{itemize}

\begin{figure}
    \centering
    \small
    \begin{tikzpicture}
    \newcommand{\bendy}{0.8}
    \newcommand{\xshift}{3cm}
    \newcommand{\yshift}{-1.5cm}
    \newcommand{\legendYshift}{-4.75cm}
    \newcommand{\legendLineLength}{1cm}
    \newcommand{\legendYNewKey}{-0.6cm}
    \newcommand{\legendValueXshift}{0.75cm}
    \newcommand{\headYShift}{1.5cm}
    \newcommand{\fontSizeModels}{\footnotesize}
    \newcommand{\legendFontSize}{\normalsize}
    \tikzset{
        header/.style={
                text centered,
                align=center,
            },
        block/.style={
                text centered,
                align=center,
                draw,
                text depth=0cm,
                rectangle,
                minimum width=1.5cm,
                font=\fontSizeModels,
            },
            incomparable/.style={
                    dashed,
                },
            incomparableW/.style={
                    dotted,
                },
        leg/.style={
                anchor=west,
                align=left,
            },
    }

    \begin{scope}[xshift=-\xshift]
        \node[header] at (0, \headYShift) {Lifted Graphs IR};
        \node[block] (ilg)    at (0, 0.5*\yshift) {ILG\\ \cite{chen.etal.2024a}};
        \node[block] (muninn) at (0, 1.5*\yshift) {Muninn\\ \cite{staahlberg.etal.2022}};
        \draw[->, ] (muninn) -- (ilg) node [midway, right] {};
    \end{scope}

    \begin{scope}[xshift=0]
        \node[header] at (0, \headYShift) {Grounded Graphs};
        \node[block] (asnets) at (0, 0) {ASNets\\ \cite{toyer.etal.2020}};
        \node[block] (slg) at (0, 1*\yshift) {SLG*\\ \cite{chen.etal.2024}};
        \node[block] (hgn) at (0, 2*\yshift) {STRIPS-HGN*\\ \cite{shen.etal.2020}};
        \draw[->,draw=red] (hgn) -- (slg) node [midway, right] {};
        \draw[->,draw=red] (slg) -- (asnets) node [midway, right] {};
    \end{scope}

    \begin{scope}[xshift=\xshift]
        \node[header] at (0, \headYShift) {Lifted Graphs PR};
        \node[block] (ploi) at (0, 0.5*\yshift) {PLOI\\ \cite{silver.etal.2021}};
        \node[block] (ob)   at (0, 1.5*\yshift) {OBS\\ \cite{horcik.sir.2024}};
        \draw[->] (ob) -- (ploi) node [midway, right] {};
    \end{scope}

    \draw[->,draw=green] (ilg.east) -- (asnets.west)  node [midway, above left] {};
    \draw[incomparableW,draw=blue,very thick] (ploi.west) -- (asnets.east) node [midway, above right] {};
    \draw[incomparable,draw=blue] (ilg) to[out=90, in=90, looseness=\bendy, edge node={node [above] {}}] (ploi);
    \draw[incomparable,draw=red] (ilg.east) -- (slg.west)  node [midway, above right] {};
    \draw[incomparable,draw=red] (ploi.west) -- (slg.east) node [midway, above left] {};

    \legendFontSize
    \begin{scope}[xshift=-\xshift-0.5cm, yshift=\legendYshift]
        \node (Xa) at (0, 0) {X};
        \node (Ya) at (\legendLineLength, 0) {Y};
        \draw[->] (Xa) -- (Ya) {};

        \node (Xb) at (0, \legendYNewKey) {X};
        \node (Yb) at (\legendLineLength, \legendYNewKey) {Y};
        \draw[incomparable] (Xb) -- (Yb) {};

        \node (Xb) at (0, 2*\legendYNewKey) {X};
        \node (Yb) at (\legendLineLength, 2*\legendYNewKey) {Y};
        \draw[dotted, very thick] (Xb) -- (Yb) {};

        \node (star) at (0.5*\legendLineLength, 3*\legendYNewKey) {X*};

        \node[leg] at (\legendLineLength + \legendValueXshift, 0) {Y is strictly more expressive than X};
        \node[leg] at (\legendLineLength + \legendValueXshift, \legendYNewKey) {X and Y are incomparable};
        \node[leg] at (\legendLineLength + \legendValueXshift, 2*\legendYNewKey) {X and Y are incomparable with  no\\$h^*$ difference requirement};
        \node[leg] at (\legendLineLength + \legendValueXshift, 3*\legendYNewKey) {X is used for multi-domain heuristics};
    \end{scope}

\end{tikzpicture}
    \caption{
        Expressivity hierarchy of graph representations of planning tasks.
        {\color{green}{Co}}{\color{red}{lo}}{\color{blue}{urs}} of edges denote the key takeaways classified in Section~\ref{ssec:hierarchy}.
    }
    \label{fig:expressivity}
\end{figure}

\section{Graph Learning Models}\label{sec:models}
Another core component of a graph learning approach for planning is the graph learning model itself.
ML techniques can generally be classified into deep learning and classical taxonomies.
Deep learning~\cite{lecun.etal.2015} pipelines automatically compute \emph{parameterised encoder} functions which convert raw input data into latent features which they deem useful for inferring outputs, generally using some form of neural network.
In contrast, classical machine learning pipelines predefine the \emph{feature extractor} for converting raw input data into feature vectors which are used by an arbitrary chosen downstream inference model, such as a regression model or decision tree.

\subsection*{Classical ML is Better Suited than Deep Learning for Planning}
The graph learning equivalent of such models include GNNs and graph kernels.
Due to the popularity of deep learning, almost all recent works in learning for planning since the introduction of ASNets employ some variant of GNN as the underlying learning model.
However, it has been shown very recently that classical ML approaches such as linear graph kernels significantly outperform GNN approaches for planning~\cite{chen.etal.2024a}, over various metrics and often by several orders of magnitude.

This approach was again motivated by the simple, yet well-known result that the WL algorithm upper bounded the expressivity of MPNNs~\cite{morris.etal.2019,xu.etal.2019}, and such an algorithm can be converted into the WL graph kernel for extracting features for graphs~\cite{shervashidze.etal.2011}.
The WL graph kernel has the same polynomial computational complexity as the typical MPNN but without the need to perform matrix operations which increases its runtime often by a significant constant factor.
Furthermore, the features can be combined with a simple linear model which lead to explainable and very fast models in terms of training and evaluation.
Given that planning is a time-sensitive task and graph learning models may be called many times during planning, such speedups from using classical ML approaches over deep learning methods can lead to greater gains.
It is also the case that generating training labels for many and large planning tasks is expensive such that one should use models with high data efficiency.

\begin{figure}
    \centering
    \newcommand{\imgSize}{0.49\columnwidth}
    \includegraphics[width=\imgSize]{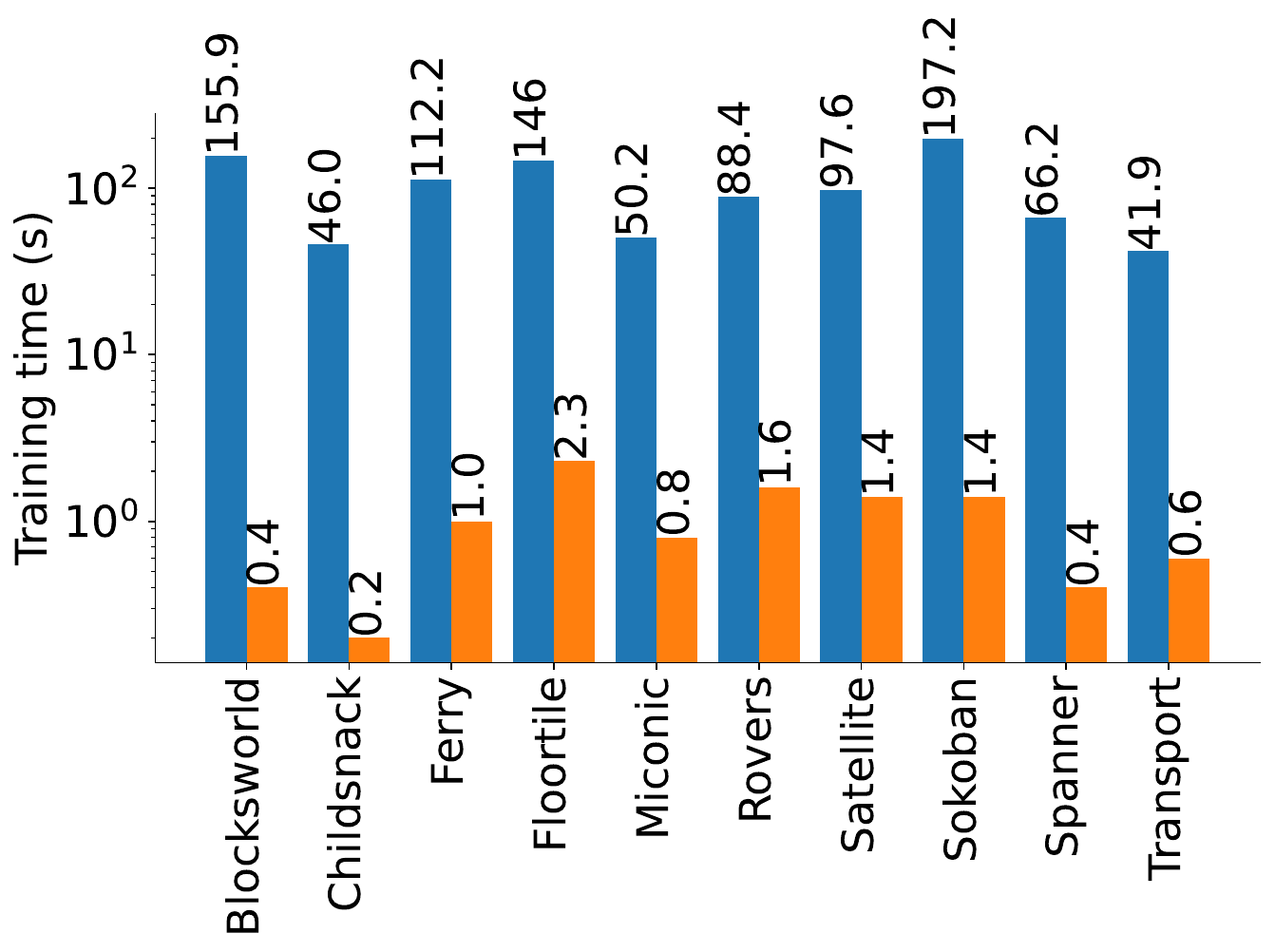}
    \includegraphics[width=\imgSize]{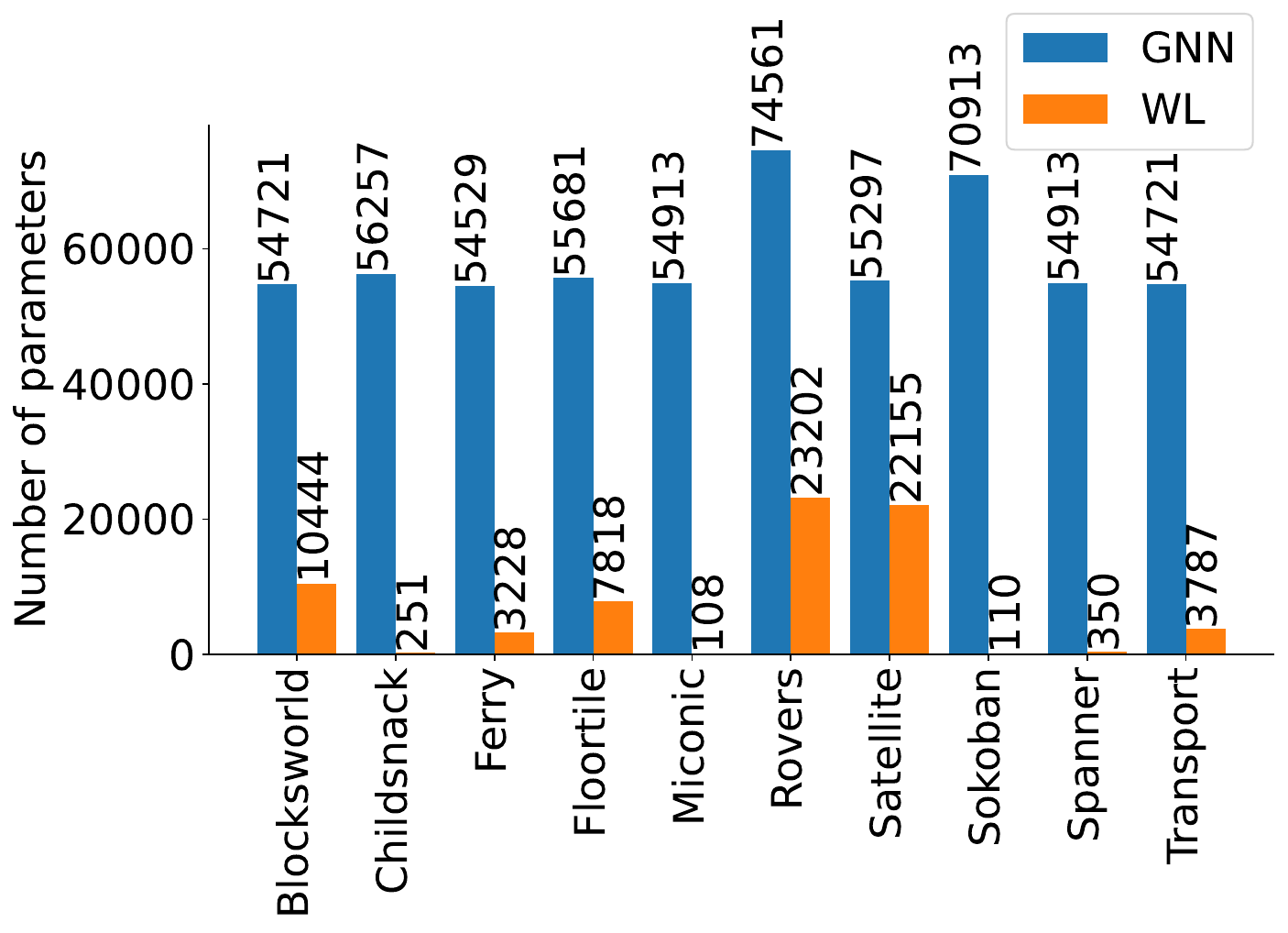}
    \captionof{figure}{Time to train after data preprocessing in log scale (left) and number of learnable parameters (right) of GNN and WL models on various planning domains.}
    \label{fig:gnn-wl-train}
\end{figure}

Indeed, Fig.~\ref{fig:gnn-wl-train} shows how linear models using WL features exhibit significantly faster training times and fewer parameters than GNN models, and Section~\ref{sec:experiments} later provides results also showcasing better planning performance.
We can further contrast this with the era of scaling with LLMs exhibiting parameters in the order of billions and training data and times so significant that they exhibit large monetary costs.
This story concerning cheap models outperforming large models still holds true in numeric planning~\cite{chen.thiebaux.2024}, where one may think neural networks may be more suited to reasoning over numbers as function approximators.

\section{Optimisation}\label{sec:optimisation}
Typical machine learning tasks have a well defined problem to solve, such as classification or regression, which usually give rise to obvious loss functions to optimise.
Conversely, the focus of planning generally involves solving problems efficiently, and in some cases optimising over solution quality.
Also in contrast to Reinforcement Learning, planning does not exhibit dense reward signals which can be used to guide learning.
Thus, there is not obvious method for deciding the optimisation problem to solve for training learners for planning.

\subsection{Learning Policies}
The seminal deep learning for planning model, ASNets~\cite{toyer.etal.2018,toyer.etal.2020}, learns policies for solving planning tasks in a RL-style fashion.
A benefit of learning policies is that when learned correctly, execution of policies is linear time in the size of the solution.
Furthermore, some planning domains only require a `trick' to be learned in order to solve them which can be represented by a neural network architecture.
One downside of learning policies is that execution of such policies have no completeness guarantees, meaning that execution of the policy may not return goal-reaching plan if it exists.
Thus, one usually combines the use of policies with search to achieve completeness~\cite{shen.etal.2019}.
Orthogonally one may allow a user to provide advice or background knowledge to the learning planner for defining the solution space for a planning domain, and instead focus the learner on \emph{optimising} solution quality, which is a more difficult task than solving domains we may already know the answer to~\cite{chen.etal.2024b}.

\subsection{Learning Optimal Heuristic Functions}
Common L4P approaches exploit the fact that most state-of-the-art planners are based on heuristic search, giving rise to the idea of learning reusable heuristic functions\footnote{Drawing analogies to RL, this is similar to learning optimal value functions across multiple tasks.} in a supervised fashion for guiding search.
Contrary to policies, learned heuristic functions used with search algorithms such as GBFS are complete and will always (eventually) return a solution if one exists.

However, most approaches end up performing the naive approach of trying to learn the optimal $h^*$ heuristic for a domain from samples via mean squared error loss.
This is generally not the best idea as theoretical arguments can be made from computational complexity theory, or explicit examples~\cite{chen.etal.2024} stating that learning $h^*$ is not possible for some domains, and furthermore, it may not even be necessary to just solve planning tasks.
This is because $h^*$ is derived from optimal solutions which are stronger than arbitrary solutions for planning tasks.
Furthermore, optimising cost-to-go as mean squared error (MSE) estimates does not match up with the original derivation of MSE loss as maximising likelihood under Gaussian distribution assumptions, which is not the case for planning heuristic values~\cite{nunezmolina.etal.2024}.

Nevertheless, the reason why researchers generally opt to learn $h^*$ is because it is a canonical label for training tasks that can be computed automatically and efficiently.
More specifically, it is easier to run an optimal planner on an arbitrary task to receive optimal plans and $h^*$ labels, than to manually label tasks with domain-dependent polynomial-time heuristic functions that still guide search efficiently.
One such example of heuristics with this property are dead-end avoiding, descending heuristics~\cite{seipp.etal.2016}.
Another reason for learning $h^*$ is that in the best case scenario where $h^*$ is learned correctly, problems can be solved in linear time with respect to solution size.

\subsection{Learning Ranking Functions}
Garrett et al.~\cite{garrett.etal.2016} proposed to frame heuristic functions not as learning cost-to-go estimates but as ranking states for expansion in GBFS.
An advantage of ranking as a learning task is that it aligns better with the intent of GBFS and furthermore allows for a larger hypothesis space of solutions.
This idea was extended to neural network architectures in independent works~\cite{chrestien.etal.2023,hao.etal.2024}.
However, the latter work identified that the data used in~\cite{chrestien.etal.2023} is quadratic in the training plans by encoding all possible ranking pairs, but can be reduced to a linear number of rankings by exploiting the fact that ranking is a transitive relation.

\subsubsection{Classification}
Ranking can be interpreted as a classification task on pairs of states.
This has been modelled and solved by RankSVMs in~\cite{garrett.etal.2016} or by backpropagation in~\cite{hao.etal.2024}.
Ranking formulations treating pairs of distinct states $(s, t)$ as single data point that can be labelled as either $1$ or $-1$, where $1$ means that $s$ is ranked strictly better than $t$, and vice versa for $-1$.
True labels are derived from optimal plan traces where $(s, t)=1$ if $t$ is either a parent or sibling of $s$ in an optimal plan trace.
After learning is performed, point-wise GBFS heuristic functions are extracted from the model that remain faithful to the ranking relation.

\subsubsection{Constrained Optimisation}
\begin{figure}
    \centering
        \trimbox{0.45cm 0.1cm 0.4cm 0.1cm}{
            \scalebox{0.825}{\begin{tikzpicture}
    \tikzset{
        outer sep=2pt,
        font=\tiny,
        state/.style={
            draw,
                circle,
                minimum width=0.45cm,
            },
        goal/.style={
            draw,
                circle,
            minimum width=0.35cm,
        },
        plan/.style={
            fill=yellow!20,
        }
    }
    \newcommand{\xshift}{1}
    \newcommand{\yshift}{0.8}
    \newcommand{\trace}{
        \node (z) at (-0.75*\xshift,0) {};
        \node[state,plan] (a) at (0,0) {};
        \node[state,plan] (b) at (1*\xshift,0) {};
        \node[state] (ba) at (1*\xshift,\yshift) {};
        \node[state] (bb) at (1*\xshift,-\yshift) {};
        \node[state,plan] (c) at (2*\xshift,0) {};
        \node[goal] (cg) at (2*\xshift,0) {};
        \node[state] (ca) at (2*\xshift,\yshift) {};
    }
    \newcommand{\segmentShift}{3.5cm}

    \begin{scope}[xshift=-\segmentShift]
        \node[rectangle, rounded corners, minimum width=3cm, minimum height=2.5cm, draw] at (0.85,0) {};
        \trace
        \node at (a) {2};
        \node at (b) {1};
        \node at (ba) {3};
        \node at (bb) {0};
        \node at (c) {0};
        \node at (ca) {-3};
        
        \draw[->] (z) -- (a);
        \draw[->] (a) -- (b);
        \draw[->] (a) -- (ba);
        \draw[->] (a) -- (bb);
        \draw[->] (b) -- (c);
        \draw[->] (b) -- (ca);
    \end{scope}

    \begin{scope}
        \node[rectangle, rounded corners, minimum width=3cm, minimum height=2.5cm, draw] at (0.85,0) {};
        \trace
        \node at (a) {6};
        \node at (b) {5};
        \node at (ba) {7};
        \node at (bb) {7};
        \node at (c) {-5};
        \node at (ca) {5};
        
        \draw[->] (z) -- (a);
        \draw[->] (a) -- (b) node[midway,above=-2pt] {$>$};
        \draw[->] (a) -- (ba);
        \draw[->] (a) -- (bb);
        \draw[->] (b) -- (c) node[midway,above=-2pt] {$>$};
        \draw[->] (b) -- (ca);

        \node at ($(ba)!0.5!(b)$) {$>$};
        \node at ($(bb)!0.5!(b)$) {$>$};
        \node at ($(ca)!0.5!(c)$) {$>$};
    \end{scope}

    \begin{scope}[xshift=\segmentShift]
        \node[rectangle, rounded corners, minimum width=3cm, minimum height=2.5cm, draw] at (0.85,0) {};
        \trace
        \node at (a) {6};
        \node at (b) {5};
        \node at (ba) {5};
        \node at (bb) {7};
        \node at (c) {-5};
        \node at (ca) {5};
        
        \draw[->] (z) -- (a);
        \draw[->] (a) -- (b) node[midway,above=-2pt] {$>$};
        \draw[->] (a) -- (ba);
        \draw[->] (a) -- (bb);
        \draw[->] (b) -- (c) node[midway,above=-2pt] {$>$};
        \draw[->] (b) -- (ca);

        \node at ($(ba)!0.5!(b)$) {$\geq$};
        \node at ($(bb)!0.5!(b)$) {$\geq$};
        \node at ($(ca)!0.5!(c)$) {$\geq$};
    \end{scope}
\end{tikzpicture}}
        }
    \caption{
        Visualisations of heuristics that achieve zero loss when optimising cost-to-go (left), classification ranking (middle), and constrained optimisation ranking (right).
        Yellow nodes correspond to optimal plan traces, and white nodes the siblings of trace states.
    }
    \label{fig:ranking}
\end{figure}
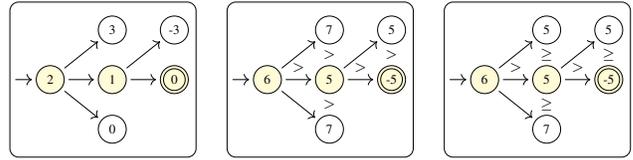

An issue with the classification formulation is that the hypothesis space is still restrictive given that cross-entropy forces ranking differences to be close to the value 1 even when this may not be necessary to represent a useful ranking function.
%
This issue can be handled by formulating the problem as a \emph{constrained optimisation} problem inspired by SVMs which allows additional flexibility in heuristic predictions as long as the rankings represented by inequalities are preserved.
Furthermore, there is also room to model the difference between comparing optimal states against its siblings which may be equally good ($\geq$) and against its parent which will always be worse with positive action costs ($>$) inequalities.
Such an example can be seen in the LP encoding in~\cite{chen.thiebaux.2024}.
Fig.~\ref{fig:ranking} illustrates the differences between learning heuristic functions as cost-to-go estimates, classification rankings, and constrained optimisation rankings.

\section{Experimental Results}\label{sec:experiments}
In this section, we provide an outline of experimental results arising from our work on graph learning for both classical (Section~\ref{secc:classic}) and numeric (Section~\ref{secc:numeric}) planning, and summarise key takeaways (Section~\ref{secc:takeaways}).

\begin{figure}
    \centering
    \includegraphics[width=0.8\columnwidth]{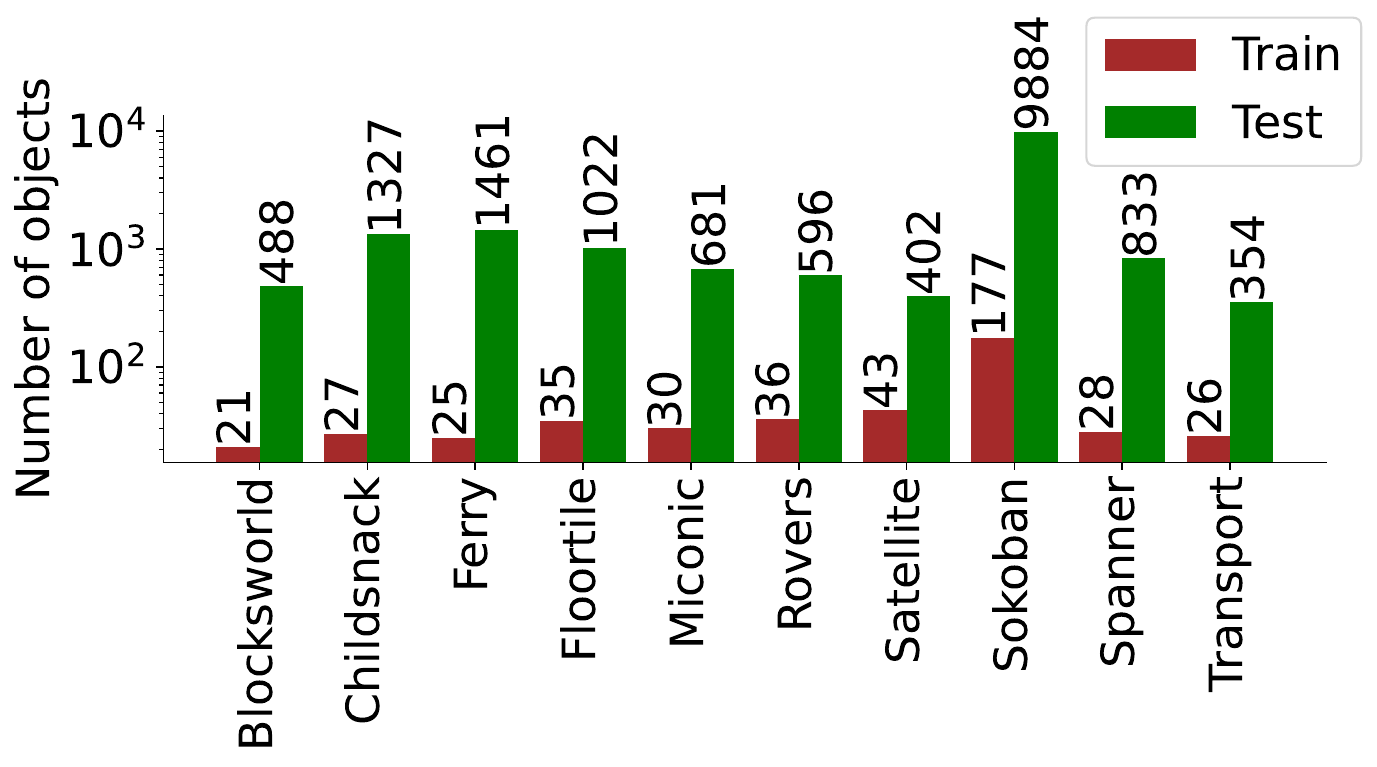}
    \caption{Sizes of training and testing tasks in the IPC23LT.}
    \label{fig:ipc23lt}
\end{figure}

\subsection{Classical Planning}\label{secc:classic}
\begin{figure}
    \centering
    \newcommand{\halfff}{0.45}
    \includegraphics[width=\halfff\columnwidth]{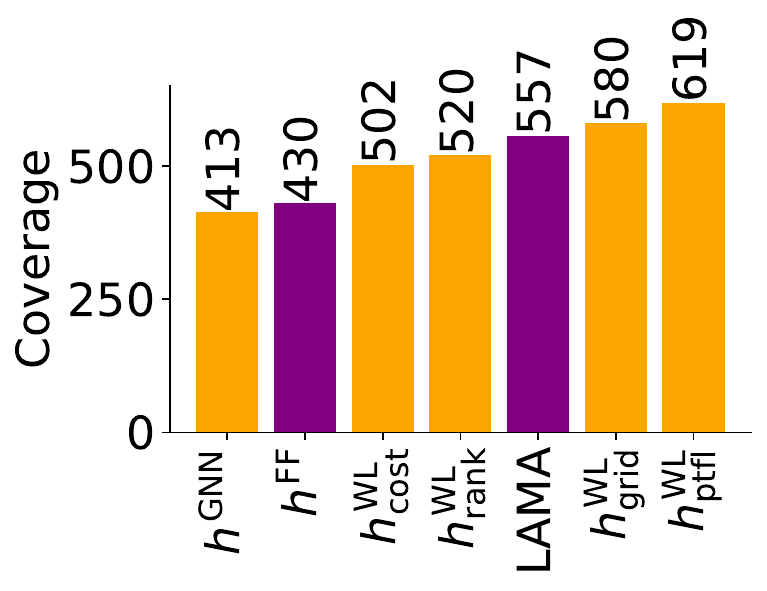}
    \raisebox{-0.135\height}{\includegraphics[width=\halfff\columnwidth]{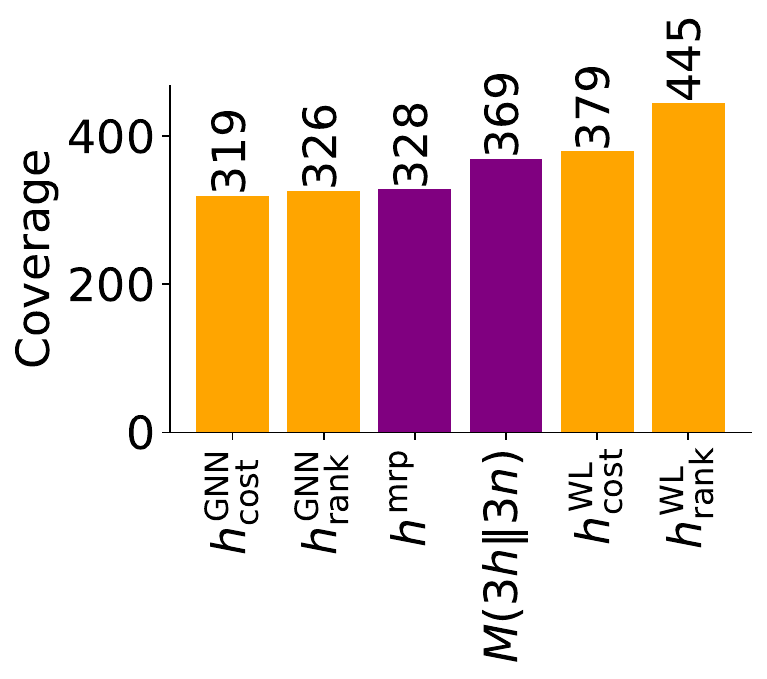}}
    \caption{
        Coverage of classic/numeric (purple) and learning (orange) planners on classical (left) and numeric (right) encodings of the IPC23LT.
        Higher values are better ($\uparrow$).
    }
    \label{fig:classic-coverage}
    \label{fig:numeric-coverage}
\end{figure}

For classical planning, we use the benchmarks from the Learning Track of the 2023 International Planning Competition (IPC23LT)~\cite{taitler.etal.2024}.
The IPC23LT benchmarks emphasise generalisation across number of objects, where as displayed in Fig.~\ref{fig:ipc23lt}, the testing tasks are often up to an order of magnitude larger than the training tasks in the number of objects.
Furthermore, sizes of planning task state spaces generally scale in a high polynomial variable with respect to the number of objects.
The primary metric we discuss in this section is the coverage of graph learning planners across all 900 problems (10 domains $\times$ 90 problems) in the IPC23LT within a fixed 30 minute time and 8GB memory limit for testing, i.e. solving a single planning task.
Fig.~\ref{fig:classic-coverage} displays the coverage of various baseline planners and graph learning planner configurations abbreviated as follows:
\begin{itemize}
    \item $h^{\text{FF}}$: GBFS + the $h^{\text{FF}}$ heuristic~\cite{hoffmann.nebel.2001}
    \item LAMA: a strong satisficing planner which uses multiple queues, helpful actions, and lazy heuristic evaluation~\cite{richter.westphal.2010}
    \item $h^{\text{GNN}}$: GBFS + a GNN heuristic trained with mean squared error loss on $h^*$ values, detailed in~\cite{chen.etal.2024a}
    \item $h^{\text{WL}}_{\text{cost}}$: GBFS + a linear WL heuristic trained with Gaussian Process Regression, detailed in~\cite{chen.etal.2024a}
    \item $h^{\text{WL}}_{\text{rank}}$: GBFS + a linear WL heuristic trained with the LP ranking formulation in~\cite{chen.thiebaux.2024}
    \item $h^{\text{WL}}_{\text{grid}}$: GBFS + a grid search over WL heuristic configurations performed on each domain
    \item $h^{\text{WL}}_{\text{ptfl}}$: GBFS + a parallel portfolio over WL heuristic configurations
\end{itemize}

\subsection{Numeric Planning}\label{secc:numeric}
In~\cite{chen.thiebaux.2024}, we also performed experiments on numeric encodings of domains from the IPC23LT, with the exception of two domains which have no benefit from numeric encodings.
Fig.~\ref{fig:numeric-coverage} displays the coverage of various baseline numeric planners and graph learning planner configurations abbreviated as follows:
\begin{itemize}
    \item $h^{\text{mrp}}$: GBFS + the $h^{\text{mrp}}$ heuristic~\cite{scala.etal.2020}
    \item $M(3h \Vert 3n)$: state-of-the-art numeric planner with multiple queues and novelty heuristics~\cite{chen.thiebaux.2024a}
    \item $h^{\text{GNN}}_{\text{cost}}$: GBFS + a GNN heuristic trained with mean squared error loss on $h^*$ values
    \item $h^{\text{GNN}}_{\text{rank}}$: GBFS + a GNN heuristic trained with a differentiable ranking loss
    \item $h^{\text{WL}}_{\text{cost}}$: GBFS + a linear WL heuristic trained with Gaussian Process Regression
    \item $h^{\text{WL}}_{\text{rank}}$: GBFS + a linear WL heuristic trained with a LP ranking formulation
\end{itemize}



\subsection{Key Takeaways}\label{secc:takeaways}
\subsubsection{Classical ML consistently outperform deep learning for symbolic planning}
From Fig.~\ref{fig:classic-coverage}, we observe that classical ML approaches using learned WL heuristics outperform their deep learning counterparts over different optimisation formulations in terms of coverage.
In classical planning, $h^{\text{WL}}_{\text{cost}}$ outperforms $h^{\text{GNN}}$ by 89 problems,
while in numeric planning, $h^{\text{WL}}_{\text{cost}}$ (resp. $h^{\text{WL}}_{\text{rank}}$) outperforms $h^{\text{GNN}}_{\text{cost}}$ (resp. $h^{\text{GNN}}_{\text{rank}}$) by 60 (resp. 119) problems.

\subsubsection{Learned ranking functions consistently outperform learned cost-to-go estimates}
From Fig.~\ref{fig:classic-coverage}, we note that learned ranking heuristics outperform their cost-to-go estimate counterparts over graph learning models in terms of coverage.
In classical planning, $h^{\text{WL}}_{\text{rank}}$ outperforms $h^{\text{WL}}_{\text{cost}}$ by 18 problems,
while in numeric planning, $h^{\text{WL}}_{\text{rank}}$ (resp. $h^{\text{GNN}}_{\text{rank}}$) outperforms $h^{\text{WL}}_{\text{cost}}$ (resp. $h^{\text{GNN}}_{\text{cost}}$) by 66 (resp. 7) problems.

\subsubsection{Learned heuristics with simple search are competitive with strong planners}
From Fig.~\ref{fig:classic-coverage} for classical planning, we observe that single instantiations of learned WL heuristics (502 and 520) with GBFS alone do not outperform the strong LAMA planner (557).
However, taking the best configuration for each domain (580), or using multiple threads for parallelised portfolios (619), learned heuristics with simple search algorithms can be competitive with planners employing stronger search algorithms.
However, for numeric planning, learned heuristics with GBFS alone (445) are a lot more competitive with numeric planners (369).

\section{Open Challenges}\label{sec:challenges}
The general L4P field is still in its infancy and poses multiple open challenges for future research.
This is reflected by the fact that there are still many domains and problems in the tested benchmarks that remain unsolved for L4P approaches.
We highlight the themes of core open challenges that we believe are crucial for advancing the state-of-the-art.

\renewcommand{\thesubsection}{\Roman{subsection}}
\renewcommand{\thesubsubsection}{Challenge \thesubsection.\arabic{subsubsection}}  

\subsection{Expressivity}
A core challenge that needs to be addressed by L4P approaches is the concept of expressivity: the ability of a model to represent solutions to planning domains.
Propositional planning is \PSPACE-complete in general~\cite{bylander.1994} but even many domains that are solvable in polynomial time cannot be solved by learning approaches~\cite{staahlberg.etal.2022,chen.etal.2024}.
Recent L4P approaches use some variant of message passing neural networks which are known to be theoretically bounded by two-variable counting logics~\cite{morris.etal.2019,xu.etal.2019,barcelo.etal.2020,grohe.2021}, with some works studying the effect of (approximate) higher-order graph learning approaches~\cite{chen.etal.2024a,staahlberg.etal.2024} and performing distinguishability tests of models~\cite{horcik.sir.2024,drexler.etal.2024a}.

However, this direction of research mirrors the graph learning literature of building more expressive models~\cite{morris.etal.2020,kriege.etal.2020,abboud.etal.2021,balcilar.etal.2021,morris.etal.2022,feng.etal.2022,zhao.etal.2022,zhao.etal.2022a,wang.etal.2023,bouritsas.etal.2023,alvarezgonzalez.etal.2024} at the cost of runtime complexity, unclear generalisation performance and unclear relevance to downstream tasks~\cite{morris.etal.2024}.
Furthermore, tractable graph learning architectures alone have yet to be able to achieve expressivity for the basic \P-time complexity time which is not directly achievable with finite-variable counting logics, and bounded number of message passing layers.

Possible directions for research in expressivity involve building or making use of models which can handle recursion such as inductive logic programming~\cite{cropper.etal.2022} or Generalised Planning~\cite{srivastava.2010,celorrio.etal.2019} approaches, motivated by the fact that \P{} is captured by first-order logic with transitive closure~\cite{vardi.1982,immerman.1982}.
Indeed, some earlier works have studied rule-based approaches for learning policies or subgoals for planning~\cite{khardon.1999,gretton.thiebaux.2004,illanes.mcilraith.2019,bonet.geffner.2024,drexler.etal.2024}, with recent work explicitly studying the use of Datalog for directly encoding planning domain solutions and solvability~\cite{grundke.etal.2024,chen.etal.2024b}.

\subsection{Generalisation}
Another key challenge in L4P is understanding both theoretical and empirical generalisation results for planning.
L4P is inherently an out-of-distribution task as testing tasks are arbitrarily large and hence are drawn from a different distribution from bounded-size training tasks.
Thus, exploiting common generalisation theory tools such as VC dimension~\cite{vapnik.1998} and Rademacher complexity~\cite{bartlett.mendelson.2001}, which assume similar training and testing probability distributions, for bounding generalisation theory is not straightforward.

Conversely, planning representations often contain information encoded as rich relational or logical structures.
Related to the concept of expressivity previously, researchers have developed theoretical frameworks to better understand the behaviour of planning domains such as novelty width~\cite{lipovetzky.geffner.2012}, correlation complexity~\cite{seipp.etal.2016}, the river measure~\cite{dold.helmert.2024}, and methods to bound such measures~\cite{dold.helmert.2024a}.
These tools may provide insights for developing generalisation theory for planning.

\subsection{Optimisation}
As discussed in Section~\ref{sec:optimisation}, there is no clear consensus on the best optimisation criteria for learning for planning as this may depend on the domain, learning architecture and also training data.
Although ranking as discussed provides a much better suited optimisation criteria for learning heuristics for planning, it may be the case that learning other forms of domain knowledge such as policies, subgoals~\cite{drexler.etal.2024}, and search effort estimates~\cite{orseau.lelis.2021,ferber.etal.2022} may be easier and well-suited for specific domains.
In other words, there is no single optimisation criterion that is best for all planning domains.
Theoretical and empirical results on optimisation criteria that are well suited for specific planning domain characteristics are still unknown.

\subsection{Collecting Data}
Another challenge in L4P is the problem of deciding how and what data to collect for training, mirroring the exploration-exploitation tradeoff in RL~\cite{sutton.barto.1998}.
This also relates to the optimisation problem as the choice of optimisation criteria depends on the type of data collected.
For example, optimal plan traces are sufficient for learning cost-to-go or ranking heuristic functions, but not optimal policies.
This is because there may be more than one optimal action to take at each state that cannot be deduced from optimal plans.
ASNets~\cite{toyer.etal.2018,toyer.etal.2020} handles this issue similar to the RL approach of exploring with a partially learned policy and exploiting with a teacher planner as proxy for a reward function.
Learning subgoals as policy sketches requires expanding entire state spaces~\cite{drexler.etal.2024} which does not scale to high predicate or asymmetric domains, even with symmetry detection~\cite{drexler.etal.2024a}.

Similarly to the optimisation problem, theoretical and empirical results on how to best collect and how much data to collect for L4P are much appreciated.

\subsection{Fair Comparisons}
A final challenge in L4P is the problem of fairly comparing different methodologies and approaches, given the additional experimental variables of training data and training time introduced when learning is involved.
Thus, this resulted in various researchers and groups employing different benchmarks and evaluation strategies.
The Learning Track of the 2023 International Planning Competition~\cite{taitler.etal.2024} was a welcome addition for standardising the set of planning domains and testing tasks.
However, although training tasks are given, the method of generating useful labels is not standardised and instead competitors were given a fixed time limit to both generate labels and train models.
This leads to an undesirable scheduling task of trading off between data generation and training time.

Possible suggestions for future proposed benchmarks or good practices to follow include (1) providing all baselines with the same set of \emph{labelled} training data generated by a fixed time limit that benefits all models, and/or (2) recording training time and amount of data used for all baselines.

\section{Conclusion}\label{sec:conclusion}
Learning for Planning (L4P) is an increasingly popular research field, with planning being one of the few problems in AI that has been unsolved by deep learning and large models~\cite{valmeekam.etal.2023,valmeekam.etal.2023a,valmeekam.etal.2024}.
This paper summarises the contributions of the authors in L4P, with a specific focus on using cheap and efficient graph learning approaches for planning.
%
%
Furthermore, we have identified 5 key challenges in L4P that still remain unsolved and offer plenty of opportunities for future research for progressing the field.

\bibliographystyle{alpha}
\bibliography{support/jair}

\end{document}